\documentclass[twoside]{article}

\usepackage[preprint]{aistats2026}
\usepackage[round]{natbib}

\usepackage{graphicx}
\usepackage{doi}
\usepackage{algorithm}
\usepackage{algpseudocode}
\usepackage{todonotes}
\usepackage{fancyhdr}
\usepackage{epstopdf} 
\usepackage{amsmath}
\usepackage{amsfonts}
\DeclareMathOperator*{\argmax}{arg\,max}
\DeclareMathOperator*{\argmin}{arg\,min}

\usepackage{amsthm}

\begin{document}

%

%

\twocolumn[

\aistatstitle{Constrained Bayesian Optimisation with Multiple Information Sources}
\aistatsauthor{
	Hauke Maathuis \And Roeland De Breuker \And Saullo G. P. Castro \And Maike Osborne
}

\aistatsaddress{
	TU Delft \And TU Delft \And TU Delft \And University of Oxford
}

]

\begin{abstract}
Bayesian Optimisation (BO) under unknown constraints is particularly challenging when feasible regions are small. In such settings, existing methods that typically rely solely on evaluations of the true objective and constraints struggle to efficiently explore the design space. However, many real-world applications offer auxiliary data sources (e.g. surrogate models or simplified simulations) that can support early exploration. Despite this potential, their integration into constrained BO remains largely unexplored. We propose a general multi-source framework that extends constrained Max-value Entropy Search, capturing inter-source correlation while balancing evaluation cost and information gain. Experiments on both synthetic and physics-based benchmarks show that our method efficiently identifies feasible and optimal solutions, even when auxiliary data are only weakly correlated. The proposed approach consistently outperforms existing methods, particularly in early-stage exploration.
\end{abstract}

\section{Introduction}\label{ch:intro}
\textit{Bayesian optimisation} (BO) is a principled framework for optimising expensive black-box functions. Such problems are prevalent in science and engineering applications, including materials discovery, drug design and simulation-based engineering. In these domains, where each evaluation may involve expensive physical experiments or high-resolution simulations, sample efficiency becomes critical. Constraints, often black-box functions themselves, are often equally expensive to evaluate, yet essential to ensure feasible designs. In many applications the feasible region is small, discontinuous, or highly non-linear, making even the discovery of a single feasible point challenging, especially in high-dimensional spaces where data is sparse, as commonly seen in crashworthiness design \citep{raponi_kriging-assisted_2019}. \\~\\
In many engineering applications, however, practitioners may have access to multiple information sources. Examples include coarse simulations \citep{wu_practical_2019}, simplified physics-based models \citep{maathuis_exploring_2024, aretz_multifidelity_2025}, or analytical approximations \citep{anand_crashworthiness_2024},  that are cheaper to evaluate but potentially biased or noisy. In drug design, this may include inexpensive simulations alongside costly laboratory experiments. While these information sources are often correlated with the target information source (ground truth), the strength and structure of this correlation can vary significantly. Yet, even when information sources are only weakly correlated with the target, they can still provide valuable information for optimisation and help populate the design space, particularly in data-sparse scenarios. \\~\\
Unlike traditional BO methods that rely on a single information source, \textit{multi-source BO} leverages multiple models in a cost-aware manner, using cheap sources to guide exploration while reserving expensive evaluations to maintain accuracy. These approaches, often referred to as \textit{multi-fidelity} in literature, were initially modelled through autoregressive processes \citep{2000Kenedy, forrester_multi-fidelity_2007}, with later work extending acquisition strategies to this setting \citep{kandasamy_multi-fidelity_2017, poloczek_multi-information_2016, wu_practical_2019, takeno_multi-fidelity_2020}. While much of the literature uses the term multi-fidelity, our work adopts the more general \textit{multi-source} perspective, which does not assume a strict fidelity hierarchy.\\~\\
Despite recent progress, the integration of multi-source modelling with constrained BO remains under-explored, as most methods address either constraints or multiple sources in isolation. In particular, leveraging low-cost sources for early-stage exploration, especially when feasible points are unknown, has received little attention. To close this gap, we propose a scalable framework for constrained BO with multiple information sources, targeting problems with expensive target evaluations, hard-to-locate feasible regions, and sparse data. We formalise the constrained optimisation problem with multiple sources as
\begin{equation}
	\label{eq:optimisation_problem}
	\begin{aligned}
		\min_{\mathbf{x} \in \mathcal{X} \subset \mathbb{R}^d} \quad & f^{(L)}(\mathbf{x}) \\
		\text{s.t.} \quad & c_i^{(L)}(\mathbf{x}) \leq 0, \quad i = 1, \ldots, g
	\end{aligned}
\end{equation}
where \( f^{(L)} :\mathcal{X} \to \mathbb{R} \) and \( c_i^{(L)}:\mathcal{X} \to \mathbb{R} \ \forall i = 1,...,g \) respectively denote the objective and constraints at the target information source $L$ (most accurate but costly). We assume that the objective and all constraints are always evaluated jointly at a given source. While this assumption may not hold in all domains, it is natural in computational engineering. The design space \( \mathcal{X} \subset \mathbb{R}^d \) may be high-dimensional. To reduce queries to source $L$, we additionally leverage cheaper information sources \( f^{(\ell)} \), \( c_i^{(\ell)} \) with \( \ell \in \mathcal{S} = \{0, 1, \ldots, L\} \) which provide biased or noisy estimates at lower cost. These auxiliary sources help explore the design space and identify promising regions, while target source evaluations refine solutions near the feasibility boundary or optima. The goal is to efficiently optimise \( f^{(L)} \) subject to the constraints \( c_i^{(L)} \leq 0 \) under a limited budget by balancing information gain and query cost across sources. Our work makes the following key contributions:
\begin{itemize}
	\item A unified framework for constrained BO with multiple sources, combining cheap auxiliary samples and a trust region heuristic for scalability in to high-dimensional input spaces.
	\item A systematic comparison of scaling multi-source GP models.
	\item Extensive experiments on real-world and synthetic benchmarks, showing clear improvements over existing constrained BO methods.
\end{itemize}

\section{Related Work}\label{ch:related}

Early approaches to \textit{multi-source} or \textit{multi-fidelity} modelling used hierarchical Gaussian processes (GPs) with nested fidelities \citep{forrester_multi-fidelity_2007, 2000Kenedy}, later extended to non-nested settings via autoregressive and discrepancy-based models \citep{gratiet_recursive_2013, perdikaris_nonlinear_2017}. In BO, \citet{kandasamy_multi-fidelity_2017} proposed a bandit-based fidelity selection method, while \citet{poloczek_multi-information_2016} introduced the \textit{Multi-Information Source Optimisation} (MISO) framework, treating each source as a biased observation of a latent function. Further developments include trace-aware knowledge gradients \citep{wu_practical_2019}, multi-fidelity MES \citep{takeno_multi-fidelity_2020}, and safeguards against uninformative sources \citep{mikkola_multi-fidelity_2022}. Additionally, recent pre-prints \citep{foumani_constrained_2025,cordelier_multi-fidelity_2025} explore constrained multi-source BO. The former introduces a simple cost-aware heuristic based on expected improvement with feasibility filtering, while the latter performs sequential optimisation over design and fidelity. However, both approaches remain limited in empirical validation.\\~\\
A parallel line of research has explored information-theoretic acquisition functions which aim to maximise expected information gain about the global optimum. \textit{Predictive Entropy Search} (PES) \citep{hernandez-lobato_predictive_2014} and \textit{Max-value Entropy Search} (MES) \citep{wang_max-value_2018} estimate information gain about the global optimum. Extensions include \textit{Fast Information-theoretic BO} (FITBO) \citep{ru_fast_2018} and \textit{General-purpose Information-based BO} (GIBBON) \citep{moss_gibbon_2021}. Constrained variants incorporate feasibility modelling via PES with constraints (PESC) \citep{hernandez-lobato_general_2016} and constrained MES (CMES) \citep{perrone_constrained_2019, takeno22a}. \\~\\
For scalability to high dimensions with constraints, proposed strategies include trust-region heuristics \citep{eriksson_scalable_2021}, its extension of feasibility-aware refinements \citep{ascia_feasibility-driven_2025}, and scaled lengthscale priors \citep{hvarfner_vanilla_2024, papenmeier_understanding_2025}, often in combination with \textit{log Constrained Expected Improvement} (logCEI) \citep{ament_unexpected_2023}. High-dimensional problems with thousands of black-box constraints have been addressed in \citet{maathuis_scaling_2025}, while \citet{om_posterior_2025} employ flow-based ensembles to improve GP scalability.\\~\\
Despite this progress, the joint treatment of constraints and multiple sources to scale BO algorithms remains under-explored. Early discovery of feasible regions is especially critical, yet existing methods rarely exploit auxiliary sources for this purpose. We address this gap by proposing a scalable, unified, information-theoretic framework for constrained BO with multiple sources.

\section{Bayesian Optimisation with Black-Box Constraints and Multiple Information Sources}

In this section, we first review BO with unknown constraints and its use of Gaussian processes. We then show how this framework can be extended to incorporate multiple information sources.

\subsection{Constrained Bayesian Optimisation}\label{ch:cbo}
We consider the constrained optimisation problem in Equation~\ref{eq:optimisation_problem}, where the feasible set under the target information source $L$ is defined as:
\begin{equation}
	\mathcal{X}_f^{(L)} = \{ \mathbf{x} \in \mathcal{X} \mid c^{(L)}_i(\mathbf{x}) \leq 0, \, i = 1, \ldots, g \}.
\end{equation}
Since constraint predictions may vary across sources, feasibility must be defined with respect to the target source $L$. BO \citep{kushner_versatile_1962,kushner_new_1964} addresses expensive, black-box problems by learning probabilistic surrogate models, typically GPs, to approximate the objective and constraints, allowing for efficient exploration and exploitation of the design space \citep{frazier_tutorial_2018}. An acquisition function \( \alpha(\mathbf{x}; \mathcal{D}_n): \mathcal{X} \to \mathbb{R} \) balances this exploration-exploitation trade-off by selecting the next evaluation point \( \mathbf{x}_+ \in \mathcal{X}\). In the constrained setting, feasibility such that  \( \mathbf{x}_+ \in \mathcal{X}_f\), must be accounted for by the acquisition function:
\begin{equation}
	\mathbf{x}_+ = \argmax_{\mathbf{x} \in \mathcal{X}_f} \alpha(\mathbf{x}; \mathcal{D}_n).
\end{equation}
Popular constrained acquisition functions have been proposed in  \citet{gardner_bayesian_2014,gelbart_bayesian_2014,ament_unexpected_2023,hernandez-lobato_general_2016,eriksson_scalable_2021}. These methods leverage GP-based uncertainty estimates to guide the search towards promising, feasible regions, enabling efficient optimisation despite limited budgets and unknown constraint landscapes.

\subsection{Gaussian Process Regression}
GPs provide a flexible, non-parametric prior over functions. A function \( u : \mathcal{X} \to \mathbb{R} \), where \( \mathcal{X} \subset \mathbb{R}^d \), is said to follow a GP prior if for any finite collection of input points \( \{\mathbf{x}_i\}_{i=1}^n \subset \mathcal{X} \). The corresponding vector of function values follows a multivariate normal distribution $u \sim \mathcal{GP}(\mu, \Sigma)$ such that 
\begin{equation}
	[u(\mathbf{x}_1), \dots, u(\mathbf{x}_n)]^\top \sim \mathcal{N}(\boldsymbol{\mu}, \mathbf{K}),
\end{equation}
where \( \mu : \mathcal{X} \to \mathbb{R} \) is the mean function with \( \mu_i = \mu(\mathbf{x}_i) \), and \( \Sigma : \mathcal{X} \times \mathcal{X} \to \mathbb{R} \) is the covariance function with \( K_{ij} = \Sigma(\mathbf{x}_i, \mathbf{x}_j) \). In practice, the mean is often taken to be zero, while typical covariance functions include the squared exponential or Matérn kernels, parametrised by hyperparameters that are obtained by optimising the marginal likelihood \citep{rasmussen_gaussian_2006}.

\subsection{Extending Gaussian Processes to Multiple Data Sources}\label{ch:miso}

\begin{figure}[H]
	\centering
	\includegraphics[width=0.5\textwidth]{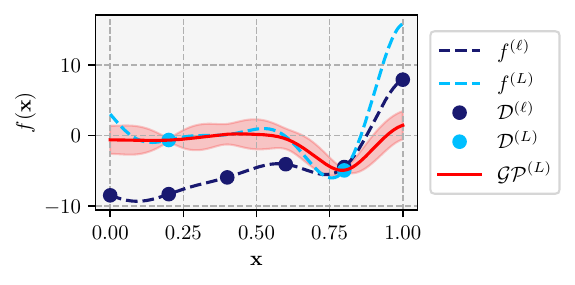}
	\caption{Multi-source GP approximation of the target function.}
	\label{fig:miso_example}
\end{figure}

To extend GPs to the multi-source setting, we model evaluations indexed by $\ell \in \mathcal{S} = \{0,1,\dots,L\}$, where $\ell = L$ denotes the most target source and $\ell = 0$ the cheapest auxiliary source. The joint domain is then the source–augmented input space $\mathcal{S} \times \mathcal{X}$, with $\mathcal{X} \subset \mathbb{R}^d$. Following the MISO framework \citep{poloczek_multi-information_2016}, we consider each source specific function \( u^{(\ell)} : \mathcal{S} \times \mathcal{X} \to \mathbb{R} \): 
\begin{equation}\label{eq:miso}
	u^{(\ell)}(\mathbf{x}) = u^{(L)}(\mathbf{x}) + \Delta^{(\ell)}(\mathbf{x}), \quad \text{with } \Delta^{(L)} \equiv 0.
\end{equation}
where \( u^{(L)}: \mathcal{X} \to \mathbb{R}\) is a latent target function and $\Delta^{(\ell)} : \mathcal{X} \to \mathbb{R}$ a source-dependent discrepancy. We place independent GP priors over both: 
\begin{equation}
	\begin{aligned}
		u^{(L)} &\sim \mathcal{GP}(\mu_L, \Sigma_L), \\
		\Delta^{(\ell)} &\sim \mathcal{GP}(\mu_\ell, \Sigma_\ell), \quad \forall\, \ell < L.
	\end{aligned}
\end{equation}
We typically assume \( \mu_\ell \equiv 0 \), whereas \( \Sigma_L \) and \( \Sigma_\ell \) are chosen from standard kernel families and learned jointly via marginal likelihood maximisation. Since the sum of independent GPs is again a GP, the combined model defines a joint GP over the source-augmented input space: $u : \mathcal{S} \times \mathcal{X} \to \mathbb{R}$ with $u(\ell, \mathbf{x}) := u^{(\ell)}(\mathbf{x}) \sim \mathcal{GP}(\mu, \Sigma)$ with mean and covariance functions given by:
\begin{equation}\label{eq:miso_mu_sigma}
	\begin{aligned}
		&\mu(\ell, \mathbf{x}) = \mathbb{E}[u^{(\ell)}(\mathbf{x})] = \mathbb{E}[u^{(L)}(\mathbf{x})] + \mathbb{E}[\Delta^{(\ell)}(\mathbf{x})], \\
		&\Sigma\big((\ell, \mathbf{x}), (\ell', \mathbf{x}')\big) = \Sigma_L(\mathbf{x}, \mathbf{x}')  + \delta(\ell,\ell') \Sigma_\ell(\mathbf{x}, \mathbf{x}'),
	\end{aligned}
\end{equation}
where \( \delta(\ell,\ell') \) is the Kronecker delta. This structure captures inter-source correlations through a shared latent process, while allowing source-specific discrepancies. Unlike naïve multi-output GPs \citep{bonilla07a} or multi-fidelity models with a strict hierarchy \citep{2000Kenedy}, the additive characteristic explicitly encodes a source structure, allowing for joint inference over the latent function \( u^{(L)} \), discrepancies \( \Delta^{(\ell)} \) and hyperparameters from observations \( \mathcal{D}_n = \{(\ell_i, \mathbf{x}_i, y_i)\}_{i=1}^n \), while maintaining flexibility. \\~\\
In practice, we fit a separate MISO model for each black-box function, i.e. one for the objective $f$ and one for each constraint $c_i \ \forall i = \{1,...,g\}$, all defined over the same source-augmented domain $\mathcal{S} \times \mathcal{X}$. Figure~\ref{fig:miso_example} illustrates how the model can approximate the target source $f^{(L)}$ from only two target source observations $\mathcal{D}^{(L)}$ by leveraging six auxiliary observations $\mathcal{D}^{(\ell)}$.

\section{Constrained Max-Value Entropy Search with Multiple Information Sources}\label{ch:mscmes_main}
We propose \textit{Multi-Source Constrained Max-value Entropy Search} (MS-CMES), extending the Max-value Entropy Search \citep{wang_max-value_2018} principle to settings where both objective and constraints can be queried at multiple information sources. In each iteration we aim to choose the next pair $(\mathbf{x}_{+}, \ell_{+})$ by balancing three factors: (i) expected information gain about the best feasible solution at the target source, (ii) the correlation between auxiliary and target sources, and (iii) the cost of each source, allowing the method to exploit cheap, approximate sources early on. This is achieved by solving
\begin{equation}\label{eq:acqf_opti_main}
	(\mathbf{x}_{+}, \ell_{+}) = \argmax_{\mathbf{x} \in \mathcal{X}, \ell \in \mathcal{S}} \frac{\alpha_n(\mathbf{x}, \ell)}{\lambda(\mathbf{x},\ell)},
\end{equation}
where $\lambda(\mathbf{x},\ell)$ denotes the cost of evaluating source $\ell$. In the following we assume that each function $f^{(\ell)}$ and constraint $c_i^{(\ell)} \ \forall i \in \{1,...,g\}$ can be observed through any source $\ell \in \mathcal{S}$.

\paragraph{Mutual Information Gain.} Following the Max-value Entropy Search principle, the function $\alpha_n$ maximises the expected information gain about the constrained optimum $f^* = \max_{\mathbf{x} \in \mathcal{X}_f^{(L)}} f^{(L)}(\mathbf{x})$ at the target source $L$:
\begin{equation}
	\alpha_n(\mathbf{x}, \ell) = \mathbb{I}\left( \mathbf{u}^{(\ell)}(\mathbf{x}); f^* \right).
\end{equation}
 with $\mathbf{u}^{(\ell)}(\mathbf{x}) = [f^{(\ell)}(\mathbf{x}), c_1^{(\ell)}(\mathbf{x}), \dots, c_g^{(\ell)}(\mathbf{x})] \in \mathbb{R}^{g+1}$. In the following we will omit that $\mathbf{u}^{(\ell)}$ depends on $\mathbf{x}$. This expression measures how much querying $(\mathbf{x}, \ell)$ reduces uncertainty over $f^*$, with all quantities modelled via the joint GP framework described in Section \ref{ch:miso}. As $f^*$ is unknown, we approximate it from the GP posteriors via: 
\begin{equation}\label{eq:f_star_main}
	f^* :=
	\begin{cases}
		\max\limits_{\mathbf{x} \in \mathcal{X}^{(L)}_f} f^{(L)}(\mathbf{x}) &\text{if } \mathcal{X}^{(L)}_f \neq \emptyset \\
		f^{(L)}\left( \argmin\limits_{\mathbf{x} \in \mathcal{X}} \sum_j \bar{c}_j^{(L)}\right) & \text{else}
	\end{cases}
\end{equation}
with $\bar{c}_j^{(L)} = \max\big(0, c_j^{(L)}(\mathbf{x}) \big)$. This fallback ensures that, in the absence of any feasible point at the target source (i.e. when $\mathcal{X}^{(L)}_f = \emptyset$), the algorithm focuses on reducing uncertainty around the most promising infeasible region. Based on this definition of $f^*$, mutual information can be written as 
\begin{equation}\label{eq:mutual_information_main}
	\begin{aligned}
		\mathbb{I}(\mathbf{u}^{(\ell)}; f^*) &= \mathbb{E}_{f^*} \left[ \mathrm{D}_\mathrm{KL}\left( p(\mathbf{u}^{(\ell)} \mid f^*) \,\|\, p(\mathbf{u}^{(\ell)}) \right) \right] \\
		&=  \mathbb{E}_{f^*} \left[ \mathbb{E}_{\mathbf{u}^{(\ell)} \mid f^*} \left[ \log \frac{p(\mathbf{u}^{(\ell)} \mid f^*)}{p(\mathbf{u}^{(\ell)})} \right] \right] 
	\end{aligned}
\end{equation}
However, \( p(\mathbf{u}^{(\ell)} \mid f^*) \) is intractable to compute directly. Therefore, we adopt the variational lower bound on mutual information as derived by \citep{takeno22a}, adapted to our multi-source context. By introducing a variational distribution $q(\mathbf{u}^{(\ell)} \mid f^*)$, we can write the information gain as:
\begin{equation}\label{eq:bounded_MI_main}
	\begin{aligned}
		\mathbb{I}(\mathbf{u}^{(\ell)}; f^*) \geq \underbrace{ \mathbb{E}_{f^*} \left[ \mathbb{E}_{\mathbf{u}^{(\ell)} \mid f^*} \log \frac{q(\mathbf{u}^{(\ell)} \mid f^*)}{p(\mathbf{u}^{(\ell)})} \right] }_{:= \alpha_n(\mathbf{x}, \ell) }.
	\end{aligned}
\end{equation}
This lower bound quantifies the expected information gain from evaluating $(\mathbf{x}, \ell)$, i.e. whether the outcome falls within the feasible region defined by the current value of the constrained optimum $f^*$.  The variational distribution $q(\mathbf{u} \mid f^*)$ is defined as the normalised posterior over the joint outputs, restricted to the feasible set $\mathcal{F} := (-\infty, f^*] \times (-\infty, 0]^g \subset \mathbb{R}^{g+1}$:
\begin{equation}\label{eq:q_set_main}
	q(\mathbf{u}^{(\ell)} \mid f^*) = 
	\begin{cases}
		\displaystyle\frac{p(\mathbf{u}^{(\ell)})}{\mathrm{Pr}(\mathbf{u}^{(\ell)} \in \mathcal{F})} & \text{if } \mathbf{u}^{(\ell)} \in \mathcal{F} \\
		0 & \text{otherwise}
	\end{cases}
\end{equation}
Substituting Equation \ref{eq:q_set_main} into the variational lower bound in Equation \ref{eq:bounded_MI_main} yields the acquisition function
\begin{equation}\label{eq:alpha_main}
	\begin{aligned}
		\alpha_n(\mathbf{x}, \ell) 
		&:= \mathbb{E}_{f^*} \left[ \mathbb{E}_{\mathbf{u}^{(\ell)} \mid f^*} \log \frac{q(\mathbf{u}^{(\ell)} \mid f^*)}{p(\mathbf{u}^{(\ell)})} \right]\\
		&\approx - \frac{1}{K} \sum_{k=1}^K \log \mathrm{Pr}(\mathbf{u}^{(\ell)} \in \mathcal{F})
	\end{aligned}
\end{equation}
where
\begin{equation}\label{eq:prob_in_F}
	\begin{split}
		\mathrm{Pr}&(\mathbf{u}^{(\ell)} \in \mathcal{F}) \\ 
		&= \mathrm{Pr}\left( f^{(\ell)}(\mathbf{x}) \leq f^* \right) \prod_{i=1}^G \mathrm{Pr}\left( c_i^{(\ell)}(\mathbf{x}) \leq 0 \right)
	\end{split}
\end{equation}
The expectation is estimated by drawing $K$ samples $\{f^*_k\}_{k=1}^K$ of the constrained optimum using discrete Thompson sampling from the GP posterior at the target source $L$. A full derivation is provided in Appendix~\ref{ch:details_derivation}.

\begin{figure}[H]
	\centering
	\includegraphics[width=0.45\textwidth]{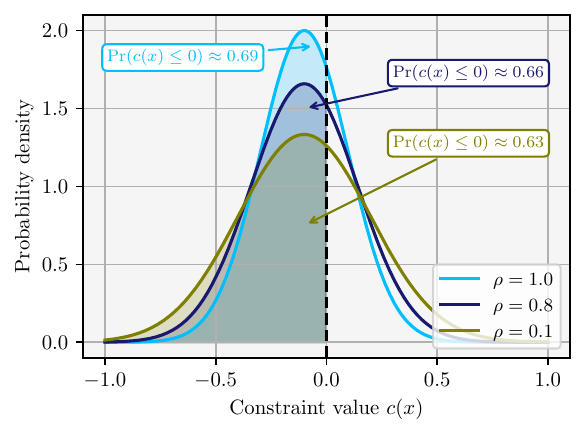}
	\caption{Illustration of the variance correction (see Equation \ref{eq:var_corr_main}), depending on the correlation coefficient $\rho \in [1.0, 0.8, 0.1] $ whereas $ \rho = 1.0 $ denotes perfect correlation.}
	\label{fig:var_correction}
\end{figure}

\paragraph{Variance correction.} As shown in Equations \ref{eq:alpha_main} and \ref{eq:prob_in_F}, we are interested in $ \mathrm{Pr}(\mathbf{u}^{(\ell)} \in \mathcal{F})$. However, this introduces a mismatch: while $\mathcal{F}$ is defined at the target source $L$, $\mathbf{u}^{(\ell)}$ is potentially computed on a different source with $\ell \neq L$. Thus, a candidate may appear feasible under source $\ell$, yet violating constraints or optimality at source $L$, since $\mathcal{X}^{(L)}_f \neq \mathcal{X}^{(\ell)}_f$. The authors in \citet{moss_mumbo_2020} show that $\mathbf{u}^{(\ell)} \mid f^{(L)}$ is not a truncated Gaussian but rather an extended skew Gaussian distribution, which they approximate in \citep{moss_mumbo_2020,moss_gibbon_2021} via a variance-correction by quantifying the correlation $\rho(\mathbf{x},\ell) \in (0,1]$ between source $\ell$ and target $L$. This mechanism ensures that weakly informative sources are automatically penalised, preventing spurious feasibility or infeasibility from dominating early-stage acquisition decisions. \\~\\
Since the MISO model assumes conditional independence between fidelities given the latent, without backward conditioning, we retain the truncated Gaussian approximation in Equation \ref{eq:alpha_main}, but correct its variance \citep{moss_gibbon_2021} via:
\begin{equation}\label{eq:sigma_tilde_main}
	\tilde{\sigma}^{(\ell)}(\mathbf{x}) \approx \sigma^{(L)}(\mathbf{x}) \left(1 - \rho^2(\mathbf{x},\ell) \Psi \left( \gamma^{(L)} \right)\right),
\end{equation}
where $\gamma^{(L)}=\frac{t-\mu^{(L)}(\mathbf{x})}{\sigma^{(L)}(\mathbf{x})}$ and $\Psi(\gamma)=\frac{\phi(\gamma)}{\Phi(\gamma)}\Big(\gamma+\frac{\phi(\gamma)}{\Phi(\gamma)}\Big)$. We set $t=f^*$ for the objective and $t=0$ for the constraints, respectively. Hence, the adjusted feasibility probabilities from Equation \ref{eq:prob_in_F} can be computed via:
\begin{equation}\label{eq:var_corr_main}
	\begin{aligned}
			\mathrm{Pr}\left( f^{(\ell)}(\mathbf{x}) \leq f^* \right) &\approx  \Phi\!\left(\frac{f^*-\mu_f^{(\ell)}(\mathbf{x})}{\tilde{\sigma}_f^{(\ell)}(\mathbf{x})}\right) \\
			 \mathrm{Pr}\left( 	c_i^{(\ell)}(\mathbf{\mathbf{x}}) \leq 0 \right) &\approx \Phi\!\left(-\frac{\mu_{c_i}^{(\ell)}(\mathbf{x})}{\tilde{\sigma}_{c_i}^{(\ell)}(\mathbf{x})}\right).
	\end{aligned}
\end{equation}
with $\Phi(\bullet)$ being the cumulative distribution function of a standard Gaussian. We emphasise that the corrected variance $\tilde{\sigma}^{(\ell)}$ is always computed with respect to the predictive mean $\mu^{(L)}$ and variance $\sigma^{(L)}$ of data source $L$, since the goal is to reduce uncertainty about the target-source optimum $f^*$.This follows from the fact that the mutual information gained by evaluating at source $\ell$ can be interpreted as upper bounded:
\begin{equation}\label{eq:I_inequ}
	\mathbb{I}(\mathbf{u}^{(\ell)}(\mathbf{x});f^*) \leq \mathbb{I}(\mathbf{u}^{(L)}(\mathbf{x});f^*),
\end{equation}
with reduction scaled by $\rho^2(\mathbf{x},\ell)$, as illustrated in Figure~\ref{fig:var_correction}.  Importantly, the scaling also prevents overconfidence of auxiliary models which may appear more certain due to being trained on a larger data set.
This correction is crucial in early-stage optimisation, where feasible regions are unknown and auxiliary sources may be misleading. By inflating variances at untrusted sources (when $\rho^2 \approx 0$), the model can still assign non-negligible feasibility probability as uncertainty is high. Rather than discarding such points, the acquisition function keeps encouraging exploration near the feasibility boundary. Details on how the correlation is computed in case of the MISO model can be found in Appendix \ref{ch:variance_correction}

\paragraph{Trust region acquisition optimisation.}
To further increase the efficiency and scalability, we restrict acquisition optimisation to a dynamically updated \emph{trust region} (TR), centred at the best observed point $\mathbf{x}^\dagger \in \mathcal{D}_n$, restricted to all points where $\ell=L$, following the idea of \citep{eriksson_scalable_2021}:
\begin{equation}
	\mathbf{x}^\dagger = \argmax_{(\mathbf{x}_i, L)\in\mathcal{D}_n} f^{(L)}(\mathbf{x}_i)
	\quad \text{s.t.} \quad c_j^{(L)}(\mathbf{x}_i)\le 0,\;\forall j ,
\end{equation}
or, if no feasible point is known, as the one with minimal total constraint violation. The TR is defined as a hypercube around $\mathbf{x}^\dagger$ with side length $r$, as
\begin{equation}
	\mathcal{T}(\mathbf{x}^\dagger, r) =
	\big\{ \mathbf{x}\in\mathcal{X}\,\big|\, 
	\mathbf{x}^\dagger-\tfrac{r}{2}\le \mathbf{x} \le \mathbf{x}^\dagger+\tfrac{r}{2}\},
\end{equation}
clipped to the domain $\mathcal{X}=[0,1]^d$. Hence, the $K$ samples $\{f^*_{k}\}_{k=1}^K$ are generated within $\mathcal{T}$, and candidate points are then selected by
\begin{equation}\label{eq:acqf_opti_main_tr}
	(\mathbf{x}_{+}, \ell_{+})
	= \argmax_{\mathbf{x}\in \mathcal{T}(\mathbf{x}^\dagger, r),\, \ell\in\mathcal{S}}
	\;\frac{\alpha_n(\mathbf{x},\ell)}{\lambda(\mathbf{x},\ell)}.
\end{equation}
Depending on the progress of the optimisation, $r$ shrinks or expands. More details can be found in Appendix~\ref{ch:TR_appendix}. A summary of the acquisition strategy is provided in Algorithm~\ref{alg:mscmes}, where the optimisation problem in Line~4 is solved using gradient ascent. The complexity is discussed in Appendix \ref{ch:complexity}.\\~\\
Summarising, the key novelty lies in combining a constrained, entropy-based information gain criterion with explicit multi-source modelling and variance correction. Unlike previous single-source constrained BO methods \citep{takeno22a} or multi-source methods without constraints \citep{poloczek_multi-information_2016}, MS-CMES provides the first principled acquisition rule for constrained optimisation that automatically down-weights uninformative sources. 

\begin{algorithm}[t]
	\caption{MS-CMES: Multi-Source Constrained Max-value Entropy Search}
	\label{alg:mscmes}
	\begin{algorithmic}[1]
		\Require Initial multi-source $\mathcal{GP}$ models for the objective and constraints, budget $B$, cost model $\lambda(x, \ell)$, Number of MC samples $K$
		\State Initialise data $\mathcal{D} \gets \{(\mathbf{x}_i, \ell_i, f^{(\ell_i)}(\mathbf{x}_i), c_j^{(\ell_i)}(\mathbf{x}_i))\}_{i=1}^n$
		\While{Computational budget is not exhausted}
		\State Sample $\{f_k^*\}_{k=1}^K$ using $\mathcal{GP}^{(L)}_{(\cdot)}$ posteriors in $\mathcal{T}$
		\State Solve $(\mathbf{x}_+, \ell_+) \gets \argmax_{\mathbf{x}\in\mathcal{T},\,\ell\in\mathcal{S}}  \frac{\alpha_n(\mathbf{x},\ell)}{\lambda(\mathbf{x},\ell)}$
		\State Query all outputs at $(\mathbf{x}_+, \ell_+)$ 
		\State $\mathcal{D}_{n+1} \gets \mathcal{D}_n \cup \{(\mathbf{x}_+, \ell_+, f^{(\ell_+)}(\mathbf{x}_+), c_j^{(\ell_+)}(\mathbf{x}_+))\}$
		\State Update GP models using $\mathcal{D}_{n+1}$
		\EndWhile
	\end{algorithmic}
\end{algorithm}

\section{Numerical Experiments}

We present numerical experiments that first compare multi-source models and their scalability with increasing dimensionality, then benchmark the proposed MS-CMES acquisition strategy against state-of-the-art methods, followed by an ablation study and parameter sensitivity analysis.

\subsection{Scalability of Gaussian Processes with Multiple Data Sources}
We compare the performance of the MISO model (Subsection~\ref{ch:miso}) against three representative alternatives: the Kennedy–O’Hagan (KOH) model \citep{2000Kenedy,forrester_multi-fidelity_2007}, multi-task Gaussian processes (MTGP) \citep{bonilla07a}, and a standard GP trained only on target-source data. As test function we use the Rosenbrock function \citep{rosenbrock_automatic_1960} in dimensions $d \in \{10,50,100\}$. The target source contains $n_L = d$ data points, while the auxiliary source contains $n_\ell = 4d$. Auxiliary signals with no, weak, and strong correlation are constructed following the procedure in Appendix~\ref{ch:mf_ext}. Figure~\ref{fig:model_comparison} reports the normalised RMSE averaged over 10 random seeds.
\begin{figure}[H]
	\centering
	\includegraphics[width=0.5\textwidth]{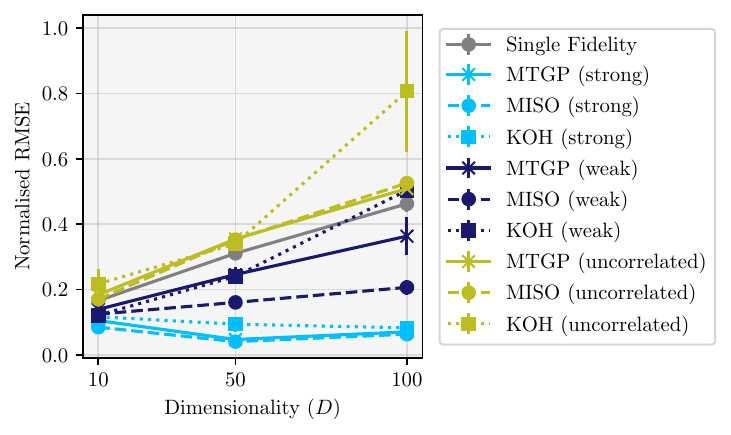}
	\caption{Model comparison across problem dimensionalities $d \in \{10, 50, 100\}$ using the Rosenbrock benchmark. Results show the normalised RMSE averaged over ten random seeds, error bars indicate $1\sigma$ standard deviation..}
	\label{fig:model_comparison}
\end{figure}
When auxiliary and target sources are strongly correlated, all multi-source models achieve comparable accuracy and consistently outperform the single-source GP. This behaviour is expected in settings such as mesh-refined aerodynamic solvers, where lower-cost simulations preserve much of the structural information of their high-cost counterparts. \\~\\
By contrast, strong correlations cannot always be assumed. In scenarios where simulators represent the same physical system but differ substantially, or where simulation data complements experimental data, the outputs are often weakly correlated or structurally divergent. In such cases, models such as KOH and MTGP, which impose a strict hierarchy or assume shared latent structure, show degraded performance. MISO, by explicitly modelling source-specific discrepancies, remains robust across a wide range of inter-source correlations.
Notably, even when auxiliary sources are entirely uninformative or uncorrelated, the performances of MISO and MTGP do not degrade much below that of a single-source GP trained only on high-fidelity data. In contrast, KOH, which assumes a strict hierarchy between sources, performs significantly worse in such settings. This robustness makes MISO particularly well-suited for general multi-source optimisation, especially when the relationships between sources are unknown, weak, or heterogeneous.

\subsection{Benchmark Tests}\label{ch:benchmark_test}

\begin{figure*}
	\centering
	\includegraphics[width=\textwidth]{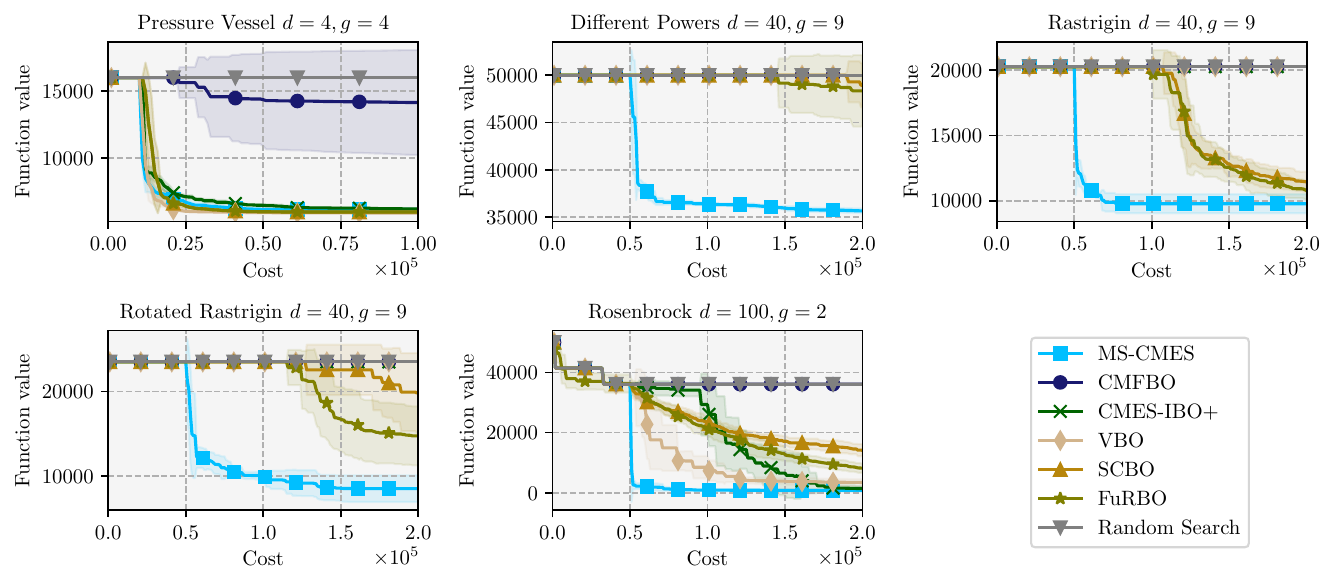}
	\caption{Comparison of optimisation performance across five constrained benchmark problems with dimensionalities up to $d=100$. Results are averaged over ten random seeds, shaded regions indicate $1\sigma$ standard deviation.}
	\label{fig:results_optimisation}
\end{figure*}

To evaluate performance and scalability, we select five constrained benchmark problems with dimensionalities ranging from $d = 4-100$. These include the physics-based Pressure Vessel benchmark \citep{coello_coello_constraint-handling_2002}, the Rosenbrock function with two constraints \citep{rosenbrock_automatic_1960}, the (Rotated) Rastrigin and the Different Powers function \citep{dufosse2022building}. More information can be found in Appendix \ref{ch:details_on_benchmarks}. We use the approach presented in Appendix \ref{ch:mf_ext} to derive a corrupted, weakly correlated signal for each objective and constraint of the respective function. Moreover, we focus on the practically relevant regime of low to moderate evaluation budgets, specifically up to 200 evaluations of the target information source, as common in scenarios where evaluation costs of the target source are immense \citep{pretsch_bayesian_2025}. The corresponding code can be found at \url{github.com/released/upon/acceptance}.
\paragraph{MS-CMES Setup.} In our MS-CMES implementation, we adopt a constant cost model, depending only on the data source, with cost weights $c_L = 1000$ for the target source and $c_\ell = 1$ for the auxiliary source (more information on the cost function can be found in Appendix \ref{ch:cost_func}). This translates to an evaluation budget of $c \approx 2 \cdot 10^5$. We use $K = 32$ samples to approximate $f^*$ and initialise the optimisation with a ratio of $n_\ell / n_L = 5$ points during the design of experiments. 
\paragraph{Baseline methods.} We compare our method against several state-of-the-art single- and multi-source approaches for constrained BO. These include SCBO \citep{eriksson_scalable_2021}, FuRBO \citep{ascia_feasibility-driven_2025}, a vanilla BO (VBO) baseline with dimensionality-scaled lengthscale priors \citep{hvarfner_vanilla_2024} using LogCEI \citep{ament_unexpected_2023} as the acquisition function, and a random search baseline. We also compare against CMES-IBO, the information-theoretic constrained BO method proposed by \citet{takeno22a}, which our work extends. In Appendix~\ref{ch:experimental_setup}, we introduce a minor modification to this method and refer to the modified version as CMES-IBO+. Additionally, we include Constrained Multi-Fidelity Bayesian Optimisation (CMFBO) from \citet{foumani_constrained_2025} as a representative fidelity-aware baseline. We do not include PESC \citep{hernandez-lobato_parallel_2017} in our comparison, as it has previously been shown to be outperformed by SCBO \citep{eriksson_scalable_2021}.
\paragraph{Results.} All GP models use Matérn kernels and are trained jointly across sources. For benchmarks with $d \geq 40$, we use a batch size of $q = 5$ and $n_0 = 50$ initial samples, while for the Pressure Vessel benchmark we use $q = 1$ and $n_0 = 10$ initial samples. For multi-source methods we use $5n_0$ auxiliary source samples. The results are depicted in Figure \ref{fig:results_optimisation}, averaged over ten random seeds with mean and variance reported. We emphasise that in multi-source methods only outputs on the target source $f^{(L)},c_1^{(L)},...,c_g^{(L)}$ determine the optimality and feasibility of a sample. Moreover, when no feasible point has been found, we assign its value to the highest feasible objective discovered so far.\\~\\
On the Pressure Vessel benchmark, all methods perform similarly, though MS-CMES achieves slightly better results than CMES-IBO+. CMFBO struggles to converge and random search fails to discover a feasible point. On the Rastrigin family (standard, rotated) and the Different Powers benchmark, most methods find it difficult to locate the first feasible point. Here, MS-CMES stands out, significantly outperforming alternatives, whereas CMFBO, CMES-IBO+, VBO, and random never find a single feasible sample. SCBO performs somewhat better but is eventually surpassed by FuRBO. On the Rosenbrock benchmark, CMFBO and random again fail to converge, while VBO and CMES-IBO+ achieve promising final values but are clearly outperformed by MS-CMES.\\~\\
Overall, MS-CMES delivers the strongest performance, particularly on the more challenging high-dimensional problems. On the Pressure Vessel benchmark, performance is broadly comparable across methods, though MS-CMES still surpasses CMES-IBO+, highlighting the benefits of incorporating auxiliary information. The advantage of our approach becomes especially evident in higher dimensions: whereas most baselines fail to identify any feasible point even after many iterations, MS-CMES discovers feasible solutions immediately after the initial design. Although the auxiliary signals are only weakly correlated (see Appendix~\ref{ch:mf_ext}), exploiting them proves highly beneficial, enabling MS-CMES to locate feasible regions early and substantially enhance optimisation performance.

\subsection{Ablation and Parameter Study}
We further examine the effect of the trust region (TR) heuristic by comparing MS-CMES with and without TR-constrained acquisition optimisation, see Figure \ref{fig:tr_ablation} in Appendix \ref{ch:ablation_and_param_study}. In the unconstrained variant, $f^*$ is sampled and Equation~\ref{eq:acqf_opti_main_tr} is optimised over the full domain $\mathcal{X}$ rather than within the adaptive TR (in $\mathcal{T}$). Both variants perform similarly on Pressure Vessel and Rosenbrock, but the TR heuristic provides clear efficiency gains on the more challenging high-dimensional problems. This indicates that dynamically restricting the search space helps concentrate exploration and prevents drifting into unpromising regions.\\~\\
We also investigate the impact of the number of Monte Carlo samples $K$ used to estimate the entropy bound. Figure \ref{fig:plot_k_param_study} in Appendix~\ref{ch:ablation_and_param_study} shows that while performance improves slightly for larger $K$, moderate values as $K=10$ are already sufficient. This underlines the robustness of the method, consistent with the findings of \citet{takeno22a}.

\section{Conclusion}
We introduced MS-CMES, the first information-theoretic framework for constrained Bayesian optimisation with multiple information sources. By explicitly modelling inter-source correlations and evaluation costs, MS-CMES leverages inexpensive, approximate sources for early exploration while reserving costly target evaluations for refinement. Across a diverse set of benchmarks, we demonstrated that MS-CMES consistently outperforms established baselines in both identifying feasible regions and converging towards optimal solutions. These results highlight the promise of multi-source modelling for constrained optimisation in domains ranging from engineering design to drug discovery.
\paragraph{Limitations.}
Our current formulation assumes that the objective and all constraints are evaluated jointly at a given source. While this assumption holds naturally in many simulation-based settings, it is restrictive in cases where partial observations are available (e.g. objective-only or constraint-only measurements). Extending MS-CMES to accommodate selective output queries would require acquisition strategies capable of balancing heterogeneous information gains. Moreover, the method currently presumes that auxiliary sources exist for every objective and constraint. In problems where this is not the case, the framework could easily be adapted by defaulting to single-source models for missing signals. Finally, performance depends on the quality of the user-specified cost function $\lambda(\mathbf{x},\ell)$. If costs are poorly estimated, the optimisation may misallocate evaluations, making careful cost modelling crucial in practice.
\paragraph{Future Work.}
Several promising directions remain. Extending the framework to support partial-output queries would increase flexibility in settings where objectives and constraints can be decoupled. In addition, scalable generative models trained on archival data offer a particularly promising avenue. We believe that by treating such models as auxiliary information sources, organisations could recycle past experiments and simulations, effectively turning historical data into a reusable asset that accelerates and guides future optimisation.

\clearpage
\bibliographystyle{unsrtnat}
\bibliography{references}  

\clearpage
\appendix
\thispagestyle{empty}

\onecolumn
\aistatstitle{Supplementary Materials}

\section{Details on MS-CMES}
This appendix complements Section~\ref{ch:mscmes_main} by providing detailed derivations for the variational bound used in MS-CMES, the associated variance correction mechanism that accounts for discrepancies between information sources and the trust region heuristic.

\subsection{Variational Bound and Variance Correction in MS-CMES}\label{ch:details_derivation}
We aim to compute the mutual information between a new observation $\mathbf{u}^{(\ell)}$ and the unknown constrained optimum $f^*$. This quantity is defined as:
\begin{equation}\label{eq:mutual_information}
	\begin{aligned}
		\mathbb{I}(\mathbf{u}^{(\ell)}; f^*) &= \mathbb{E}_{f^*} \left[ \mathrm{D}_\mathrm{KL}\left( p(\mathbf{u}^{(\ell)} \mid f^*) \,\|\, p(\mathbf{u}^{(\ell)}) \right) \right] \\
		&=  \mathbb{E}_{f^*} \underbrace{
			\left[ 
			\mathbb{E}_{\mathbf{u}^{(\ell)} \mid f^*} \left[ \log \frac{p(\mathbf{u}^{(\ell)} \mid f^*)}{p(\mathbf{u}^{(\ell)})} \right] 
			\right]	}_{(*)} 
	\end{aligned}
\end{equation}
This follows from the standard mutual information identity $\mathbb{I}(X; Y) = \mathbb{E}_{Y} \left[ \mathrm{D}_\mathrm{KL}\left( p(X \mid Y) \,\|\, p(X) \right) \right]$ with $\mathrm{D}_\mathrm{KL}$ being the Kullback-Leibler divergence. However, \( p(\mathbf{u}^{(\ell)} \mid f^*) \) is intractable to compute directly. We continue to follow the approach of \citet{takeno22a} who introduce a variational distribution $q(\mathbf{u}^{(\ell)} \mid f^*)$:
\begin{equation}\label{eq:bounded_term}
	\begin{aligned}
		\mathbb{E}_{\mathbf{u}^{(\ell)} \mid f^*} \left[ \log \frac{p(\mathbf{u}^{(\ell)} \mid f^*)}{p(\mathbf{u}^{(\ell)})} \right]  &= \mathbb{E}_{\mathbf{u}^{(\ell)} \mid f^*} \left[ \log \frac{q(\mathbf{u}^{(\ell)} \mid f^*)}{p(\mathbf{u}^{(\ell)})} \right] +\mathrm{D}_\mathrm{KL} \left( q(\mathbf{u}^{(\ell)} \mid f^*) \,\|\, p(\mathbf{u}^{(\ell)} \mid f^*) \right) \\
		&\geq \mathbb{E}_{\mathbf{u}^{(\ell)} \mid f^*} \left[ \log \frac{q(\mathbf{u}^{(\ell)} \mid f^*)}{p(\mathbf{u}^{(\ell)})} \right]
	\end{aligned}
\end{equation}
Since $\mathrm{D}_\mathrm{KL}(\cdot \,\|\, \cdot) \geq 0$ we end up at the information source-dependent lower bound. Inserting Equation \ref{eq:bounded_term} into Equation \ref{eq:mutual_information}, we can write:
\begin{equation}\label{eq:bounded_MI}
	\begin{aligned}
		\mathbb{I}(\mathbf{u}^{(\ell)}; f^*) &\geq \mathbb{E}_{f^*} \left[ \mathbb{E}_{\mathbf{u}^{(\ell)} \mid f^*} \log \frac{q(\mathbf{u}^{(\ell)} \mid f^*)}{p(\mathbf{u}^{(\ell)}(\mathbf{x}))} \right]
	\end{aligned}
\end{equation}
This lower bound quantifies the expected information gain obtained by observing whether the evaluation $(\mathbf{x}, \ell)$ yields a realisation within the feasible region associated with the current value of the constrained optimum $f^*$. In the unconstrained case, the variational distribution $q(\mathbf{u} \mid f^*)$ corresponds to a truncated Gaussian. In the constrained setting, we define $q(\mathbf{u} \mid f^*)$ as the normalised posterior over the joint outputs, restricted to the feasible set $\mathcal{F} \subset \mathbb{R}^{g+1}$, i.e. $\mathcal{F} := (-\infty, f^*] \times (-\infty, 0]^g$. We define the feasibility probability $ \mathrm{Pr}(\mathbf{u}^{(\ell)} \in \mathcal{F})$ at data source $\ell$ as:
\begin{equation}\label{eq:pr_feas}
	\begin{aligned}
 		\mathrm{Pr}(\mathbf{u}^{(\ell)} \in \mathcal{F}) = \mathrm{Pr}\left( f^{(\ell)}(\mathbf{x}) \leq f^* \right) \prod_{i=1}^G \mathrm{Pr}\left( c_i^{(\ell)}(\mathbf{x}) \leq 0 \right) \\
	\end{aligned}
\end{equation}
where $\Phi(\bullet)$ is the commulative distribution function of a standard Gaussian. Using this, the variational distribution $q(\mathbf{u}^{(\ell)} \mid f^*)$ is defined as
\begin{equation}\label{eq:q_set}
	q(\mathbf{u}^{(\ell)} \mid f^*) = 
	\begin{cases}
		\displaystyle\frac{p(\mathbf{u}^{(\ell)})}{ \mathrm{Pr}(\mathbf{u}^{(\ell)} \in \mathcal{F})} & \text{if } \mathbf{u}^{(\ell)} \in \mathcal{F} \\
		0 & \text{otherwise}
	\end{cases}
\end{equation}
Substituting Equation \ref{eq:q_set} into the lower bound in Equation \ref{eq:bounded_MI} gives: 
\begin{equation}\label{eq:alpha}
	\begin{aligned}
		\alpha_n(\mathbf{x}, \ell) 
		&:= \mathbb{E}_{f^*} \left[ \mathbb{E}_{\mathbf{u}^{(\ell)} \mid f^*} \log \frac{q(\mathbf{u}^{(\ell)} \mid f^*)}{p(\mathbf{u}^{(\ell)})} \right]\\
		&= \mathbb{E}_{f^*} \left[ \int p(\mathbf{u}^{(\ell)} | f^*) \log \frac{q(\mathbf{u}^{(\ell)} | f^*)}{p(\mathbf{u}^{(\ell)})} \, d\mathbf{u}^{(\ell)} \right]\\
		&= \mathbb{E}_{f^*} \left[ \int p(\mathbf{u}^{(\ell)} | f^*) \log \frac{p(\mathbf{u}^{(\ell)})}{ \mathrm{Pr}(\mathbf{u}^{(\ell)} \in \mathcal{F}) p(\mathbf{u}^{(\ell)})} \, d\mathbf{u}^{(\ell)} \right] \\
		&= \mathbb{E}_{f^*} \left[  \int p(\mathbf{u}^{(\ell)} | f^*) \log \frac{1}{ \mathrm{Pr}(\mathbf{u}^{(\ell)} \in \mathcal{F})} \, d\mathbf{u}^{(\ell)} \right] \\
		&= \mathbb{E}_{f^*} \left[  - \log  \mathrm{Pr}(\mathbf{u}^{(\ell)} \in \mathcal{F})  \int p(\mathbf{u}^{(\ell)} | f^*)\, d\mathbf{u}^{(\ell)} \right] \\
		&= \mathbb{E}_{f^*} \left[ - \log  \mathrm{Pr}(\mathbf{u}^{(\ell)} \in \mathcal{F}) \right] 
	\end{aligned}
\end{equation}
Finally, the expected value is approximated with $K$ samples as 
\begin{equation}
	\mathbb{E}_{f^*} \left[ - \log  \mathrm{Pr}(\mathbf{u}^{(\ell)} \in \mathcal{F}) \right] \approx - \frac{1}{K} \sum_{k=1}^K \log  \mathrm{Pr}(\mathbf{u}^{(\ell)} \in \mathcal{F})
\end{equation}

\subsection{Variance Correction}\label{ch:variance_correction}
Recalling the setup of the multi-source model in Equations \ref{eq:miso} and \ref{eq:miso_mu_sigma}, we can quantify the correlation \citep{rasmussen_gaussian_2006} with:
\begin{equation}
	\begin{aligned}
		\rho(\mathbf{x}, \ell) 
		&= \frac{\mathrm{Cov}\left(u^{(L)}(\mathbf{x}), u^{(\ell)}(\mathbf{x})\right)}{\sigma_L(\mathbf{x}) \sigma_\ell(\mathbf{x})} \\
		&= \frac{\sigma_L(\mathbf{x})}{\sigma_L(\mathbf{x}) + \sigma_\Delta(\mathbf{x})} \in (0,1].
	\end{aligned}
\end{equation} 
with $\mathrm{Cov}\!\big(u^{(\ell)}(\mathbf{x}),u^{(L)}(\mathbf{x})\big)=\mathrm{Var}\!\big(u^{(L)}(\mathbf{x})\big)={\sigma_{L}}^2$. From that, it follows that $\rho(\mathbf{x},L) \equiv 1$. The discrepancy $\Delta^{(\ell)}$ increases predictive variance at more inaccurate data sources which changes the shape of the conditional distribution. 

\subsection{Trust-Region Constrained Optimisation of the Acquisition Function}\label{ch:TR_appendix}
We constrain the optimisation of the acquisition function to a dynamically updated \emph{trust region} (TR) in order to stabilise the search. This idea, inspired by \citet{eriksson_scalable_2021}, focuses optimisation on promising neighbourhoods around the current best observed solution, which is particularly beneficial in high-dimensional settings. \\~\\
At iteration $n$, let the dataset of all evaluations be $\mathcal{D}_n = \{(\mathbf{x}_i, \ell_i, f^{(\ell_i)}(\mathbf{x}_i), c^{(\ell_i)}_1(\mathbf{x}_i), \dots, c^{(\ell_i)}_g(\mathbf{x}_i)) \}_{i=1}^n $. We select $\mathbf{x}^\dagger$ from the subset of points evaluated at the target source $L$, i.e.\ $\{(\mathbf{x}_i, \ell_i) \in \mathcal{D}_n \mid \ell_i = L\}$. If at least one feasible point has been observed, $\mathbf{x}^\dagger$ is defined as the best feasible point:
\begin{equation}
	\mathbf{x}^\dagger 
	= \argmax_{(\mathbf{x}_i, L) \in \mathcal{D}_n} 
	\; f^{(L)}(\mathbf{x}_i)
	\quad \text{s.t.} \quad c_j^{(L)}(\mathbf{x}_i) \leq 0 \ \forall j .
\end{equation}
If no feasible point exists, $\mathbf{x}^*$ is chosen as the infeasible point with the smallest total constraint violation:
\begin{equation}
	\mathbf{x}^* 
	= \argmin_{(\mathbf{x}_i, \ell_i) \in \mathcal{D}_n,\ \ell_i = L} 
	\; \sum_{j=1}^g \max \!\big(0, c_j^{(L)}(\mathbf{x}_i)\big).
\end{equation}
The trust region is defined as a hypercube centred at $\mathbf{x}^*$ with side length $r$:
\begin{equation}
	\mathcal{T}(\mathbf{x}^\dagger, r) =
	\Big\{ \mathbf{x} \in \mathcal{X} \,\big|\, 
	x^\dagger_z - \tfrac{r}{2} \leq x_z \leq x^\dagger_z + \tfrac{r}{2}, \; z=1,\dots,d \Big\},
\end{equation}
clipped to the global domain $\mathcal{X}=[0,1]^d$. \\~\\
The side length $r$ is adapted dynamically. Let $s_n$ and $f_n$ denote counters of consecutive successes and failures, respectively. A \emph{success} occurs if the new evaluation improves upon the best feasible objective, or reduces total constraint violation relative to $\mathbf{x}^*$. Otherwise, a \emph{failure} is recorded. When $s_n$ reaches a tolerance $s_{\max}$, the region is expanded as $r \gets \min(2r, r_{\max})$ and $s_n$ is reset. Conversely, if $f_n$ reaches $f_{\max}$, the region is shrunk as $r \gets \tfrac{1}{2} r$ and $f_n$ reset. If $r$ falls below $r_{\min}$, the TR is restarted around the current best point. \\~\\
At each iteration, new candidates are chosen by optimising the cost-scaled acquisition function within the TR:
\begin{equation}
	(\mathbf{x}_{+}, \ell_{+})
	= \argmax_{\mathbf{x} \in \mathcal{T}(\mathbf{x}^\dagger, r), \, \ell \in \mathcal{S}}
	\; \frac{\alpha_n(\mathbf{x}, \ell)}{\lambda(\mathbf{x},\ell)} .
\end{equation}
This strategy prevents the optimiser from wasting evaluations in unpromising regions (particularly when auxiliary sources are misleading) and provides a principled balance between local refinement and global exploration through adaptive expansion and contraction of the trust region.

\section{Computational Complexity}\label{ch:complexity}
We briefly analyse the computational complexity of the proposed method, separating the cost of model training from that of acquisition optimisation.
\paragraph{Multi-Source Gaussian Process Model} Let $n = \sum_{\ell \in \mathcal{S}} n_{\ell}$ denote the total number of observations across sources. The training and inference requires a Cholesky factorisation of the $n \times n$ covariance matrix, leading to the classical complexities for time and memory, $\mathcal{O}(n^3)$ and $\mathcal{O}(n^2)$. Optimisation of the hyperparameters requires repeated gradient evaluations, hence scales identically. Due to the additive structure, additional kernel parameters are introduced due to the discrepancy GP $\Delta^{(\ell)}$, compared to a standard, single-source GP. For large data sets $n$, inducing point methods such as Structured Kernel Interpolation (SKI) or Sparse GPs (SGPR) could be applicable. 
\paragraph{Acquisition function (MS-CMES)} The cost of evaluating the acquisition function decomposes as follows: We draw $K$ samples from the posterior of the constrained optimum on a candidate set of size $n_c$ to obtain $f^*$. This costs $\mathcal{O}(K n_c)$ plus the cost of GP posterior evaluation on $n_c$ candidates. Evaluating $\alpha(\mathbf{x},\ell)$, for each candidate $\mathbf{x}$, computing feasibility probabilities involves Gaussian CDFs of the posterior mean/variance. With $g$ constraints, a batch of $b$ candidates and  $|\mathcal{S}|$ number of sources, the complexity yields $\mathcal{O}(b K (g+1)|\mathcal{S}|)$. In total this leads to $\mathcal{O}(K n_c + b K (g+1)|\mathcal{S}|)$.
\paragraph{Computational resources in this work.}
All experiments were carried out on a compute cluster equipped with Intel Xeon Gold 5218 CPUs (2.3 GHz). All jobs were executed on single CPU nodes without GPU acceleration. Runtime per experiment varied with problem dimensionality and evaluation budget but remained within a few hours for the largest benchmarks, showing that the method does not need extensive compute. 

\section{Details on Implementation of Models}

\subsection{MISO Model}\label{ch:miso_appendix}
We implement a variant of the Multi-Information Source Optimisation (MISO) model based on the structure proposed by \citet{poloczek_multi-information_2016}. This subsection accompanies the mathematical description in Section \ref{ch:miso}. Our model assumes a hierarchical design of experiments, wherein each target source evaluation is accompanied by evaluations from all auxiliary sources at the same input location. This nested design is motivated by the assumption that auxiliary sources are inexpensive to query, allowing evaluations to be collected “for free” whenever a target source point is selected by the acquisition function. The model decomposes the objective into a shared latent function and source-specific discrepancies. Both the base kernel and the discrepancy kernels are instantiated as Matérn-5/2 kernels. Discrepancy kernels are masked to activate only on data from their associated source level. These components are combined additively in the covariance module. To capture the magnitude of inter-fidelity variation, we place a log-normal prior on the outputscale of each discrepancy kernel. The prior’s mean is set to the empirical mean squared discrepancy between target and auxiliary source observations across the training data, extracted from the nested design. This reflects an informed prior belief on the relative strength of the discrepancy signal. Finally, the model exposes a utility function that returns the local squared correlation $\rho^2(x)$ between each auxiliary source and the target source, based on the kernel structure. We implemented the model with the help of \textsc{GPyTorch} \citep{gardner_gpytorch_2021} and use it within \textsc{BoTorch} \citep{balandat2020botorch} for optimisation.

\subsection{KOH Model}
The Kennedy-O’Hagan (KOH) model \citep{2000Kenedy} is a seminal framework for multi-fidelity/ multi-source GP modelling, originally proposed for calibrating computer models using experimental data. It has since become a cornerstone in multi-source BO and surrogate modelling, particularly when relating a cheap but approximate simulator to a more expensive high-fidelity function.\\~\\
In its canonical form, the KOH model assumes a hierarchical structure that expresses the high-fidelity response $f_{L}(\mathbf{x})$ as a scaled version of the low-fidelity model $f_{\ell}(\mathbf{x})$, plus a discrepancy term:
\begin{equation}
	f_{L}(\mathbf{x}) = \rho f_{\ell}(\mathbf{x}) + \Delta(\mathbf{x}),
\end{equation}
where $\rho \in \mathbb{R}$ is a scalar calibration or scaling parameter that is learned alongside the other hyperparameters defining the model, $f_{\ell}(\mathbf{x}) \sim \mathcal{GP}(0, k_{\ell}(\mathbf{x}, \mathbf{x}’))$ is a GP representing the low-fidelity simulator, $\Delta(\mathbf{x}) \sim \mathcal{GP}(0, k_{\Delta}(\mathbf{x}, \mathbf{x}’))$ is a discrepancy GP capturing the systematic mismatch between fidelities. Under this model, the joint prior over $f_{\ell}(\cdot)$ and $f_{L}(\cdot)$ is analytically tractable and Gaussian, and the posterior predictions retain closed-form expressions under Gaussian noise. This hierarchical design enforces that high-fidelity predictions are informed by low-fidelity evaluations through the shared term $f_{\ell}(\cdot)$, while allowing the discrepancy GP to correct for systematic errors. We implemented this model with the help of \textsc{GPyTorch} \citep{gardner_gpytorch_2021}, using the Matérn-5/2 kernel.

\subsection{MTGP Model}
The MTGP model proposed by \citet{bonilla07a} offers a principled approach for jointly modelling multiple related outputs (tasks) using a shared GP framework. In the context of multi-source modelling, tasks correspond to different sources, providing a flexible and non-hierarchical alternative to models like KOH. \\~\\
Let $f_\ell(\mathbf{x})$ denote the output of task $\ell \in \{1, \dots, L\}$ at input $\mathbf{x} \in \mathcal{X}$. The MTGP assumes that the joint function $f(\mathbf{x}, \ell)$ is drawn from a GP:
\begin{equation}
	f(\mathbf{x}, \ell) \sim \mathcal{GP}(0, k((\mathbf{x}, \ell), (\mathbf{x}’, \ell’))),
\end{equation}
with a covariance function decomposed as:
\begin{equation}
	k((\mathbf{x}, \ell), (\mathbf{x}’, \ell’)) = k_{x}(\mathbf{x}, \mathbf{x}’) \cdot k_{\ell}(\ell, \ell’).
\end{equation}
Here $k_{x}$ is a standard kernel (e.g. RBF or Matérn) defined over the input space, $k_{\ell}$ is a positive semi-definite covariance matrix $\mathbf{K}_{\ell} \in \mathbb{R}^{L \times L}$ that encodes task relationships.\\~\\
This structure allows the MTGP to capture correlations across tasks (sources), enabling knowledge transfer from low-fidelity data to improve predictions at high fidelity. Importantly, unlike the KOH model, MTGPs do not impose any explicit hierarchy or discrepancy structure. Instead, task dependencies are entirely captured through the learned task covariance matrix. In this work, we make use of \textsc{GPyTorch}'s \citep{gardner_gpytorch_2021} Multi-Task Gaussian Process implementation \citep{bonilla07a}, again using the Matérn-5/2 kernel.

\section{Extending Benchmarks to Multi-Fidelities}\label{ch:mf_ext}
Let \( u^{(L)}: \mathbb{R}^d \to \mathbb{R} \) denote a target source objective or constraint function. We define a auxiliary source $u^{(\ell)}$ by applying an input-dependent oscillatory distortion to the function $u^{(L)}$. The distortion is scaled relative to the estimated magnitude of each function. Let \( S_f > 0 \) denote an empirical estimate of the objective's scale:
\begin{equation}
	S := \mathbb{E}_{x \sim \mathcal{U}([0,1]^d)} \left[ |u^{(L)}(x)| \right]
\end{equation}
The oscillatory signal is defined as:
\begin{equation}
	s(x) := \sin\left(\frac{2\pi}{d} \sum_{j=1}^{d} x_j \right)
\end{equation}
For the weak correlation level, set the relative distortion factor to $\rho := 1$, for a strong correlation $\rho := 0.1$. We define the auxiliary source \( \tilde{u}^{(\ell)} \) as:
\begin{equation}
	u^{(\ell)}(x) := u^{(L)}(x) \cdot \left( 1 + \frac{\rho S}{\max\left( | u^{(L)}(x)|, \varepsilon \right)} \cdot s(x) \right) + \xi
\end{equation}
with $\xi \sim \mathcal{N}(0,\sigma_\xi^2)$.
\begin{figure}[H]
	\centering
	\includegraphics[width=0.3\textwidth]{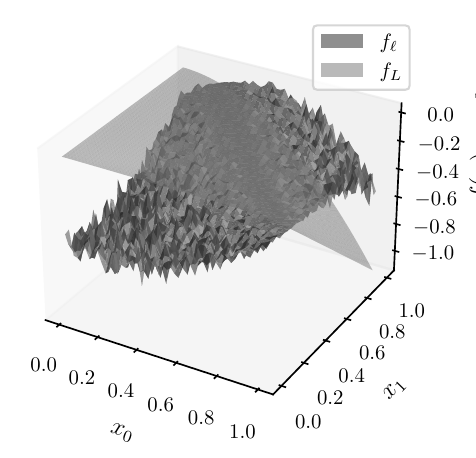}
	\caption{We use the $40$-dimensional Different Powers objective function $f$ and plot a slice along variable $x_4$ to visualise the difference between the target information source $f_L$ and the weakly correlated and noisy auxiliary information source $f_\ell$.}
	\label{fig:fig2}
\end{figure}

\section{Baseline Methods: Experiment Setup}\label{ch:experimental_setup}
\begin{figure}[H]
	\centering
	\includegraphics[width=0.7\textwidth]{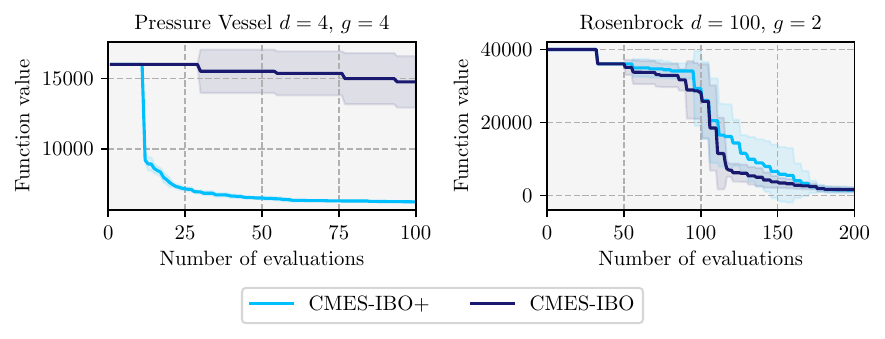}
	\caption{Comparison of the original definition of $f^*$ (CMES-IBO) verus using the definition in Equation \ref{eq:f_star_main}, here denoted with CMES-IBO+.}
	\label{fig:cmes_test}
\end{figure}

\paragraph{Multi-Source Constrained Max-value Entropy Search} In MS-CMES we use the MISO model as discussed in Appendix \ref{ch:miso_appendix} and employ discrete Thompson sampling to sample $f^*$. Moreover, we use \textsc{optimize\_acqf\_mixed} in \textsc{BoTorch} \citet{balandat2020botorch}, using 3 restarts and 200 raw samples to maximise the acquisition function.
\paragraph{Vanilla Bayesian Optimisation} Here, we use the dimensionality-scaled lengthscale prior from \citet{hvarfner_vanilla_2024} and the logarithmic version of Constrained Expected Improvements (logCEI), proposed in \citet{ament_unexpected_2023}. We use 64 MC samples and enable the \textsc{sample\_around\_best} option, embedded in \textsc{BoTorch}'s acquisition function optimiser. More information can be found in \citet{papenmeier_understanding_2025}.
\paragraph{Scalable Constrained Bayesian Optimisation} We use the same parameters as in \citet{eriksson_scalable_2021}. The method is implemented in \textsc{BoTorch} and can be found here:  \url{https://botorch.org/docs/tutorials/scalable_constrained_bo/}
\paragraph{Feasibility-Driven Trust Region Bayesian Optimisation} FuRBO is directly built upon SCBO and differs only w.r.t. the trust region heuristic. Here again, we employ the same user-defined parameters as presented in \citet{ascia_feasibility-driven_2025}. The corresponding code was taken from \url{https://anonymous.4open.science/r/FuRBO}.
\paragraph{Constrained Max-value Entropy Search} Additionally, we compare against the original CMES-IBO methods, proposed in \citet{takeno22a}. While the authors propose to define $f^*$ as 
\begin{equation}
	f^* :=
	\begin{cases}
		\max\limits_{\mathbf{x} \in \mathcal{X}_f} f(\mathbf{x}) &\text{if } \mathcal{X}_f \neq \emptyset \\
		-\infty & \text{else},
	\end{cases}
\end{equation}
we found that our definition from Equation \ref{eq:f_star_main} yields better results as depicted in Figure \ref{fig:cmes_test} where we plot the mean and standard deviation over ten runs. For all benchmarks (this Section as well as Section \ref{ch:benchmark_test}) we use $K=32$. Similar to before, we employ discrete Thompson sampling to sample $f^*$ and use \textsc{optimize\_acqf} in \textsc{BoTorch} \citet{balandat2020botorch}, using 3 restarts and 200 raw samples. 
\paragraph{Constrained Multi-Fidelity Bayesian Optimisation}
We implemented our own version of the acquisition function in \citet{foumani_constrained_2025} in combination with the introduced MISO model. We use gradient-ascent to maximise the acquisition function. However, to obtain a similar ratio of auxiliary and target source evaluations we noted that we need to set the cost to $c_L = 0$. Like before, we employ \textsc{optimize\_acqf} in \textsc{BoTorch} \citet{balandat2020botorch}, using 3 restarts and 200 raw samples. 

\section{Details on Cost Function}\label{ch:cost_func}
\begin{figure}[ht!]
	\centering
	\includegraphics[width=0.35\textwidth]{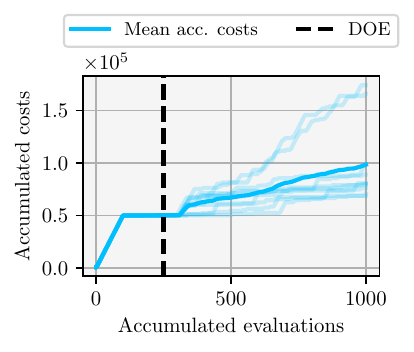}
	\caption{Accumulated evaluation cost as a function of the accumulated number of evaluations.}
	\label{fig:when_which_source}
\end{figure}

As the cost function needs to be set up such that it reflects the magnitude of the acquisition function scale, we found that in MS-CMES $\lambda(\mathbf{x},\ell) = 1 + \frac{\ell}{10^{5}} c_\ell$ balances the number of target and auxiliary source evaluations well. However, for real world problems, the choice of the cost function can be problem specific and needs to reflect the sources it is trying to balance. We emphasise that the choice of cost function is arbitrary as it solely scales the utility value of the acquisition function.\\~\\
To showcase when a target source evaluation is queried, we depict in Figure \ref{fig:when_which_source} the accumulated costs over the accumulated evaluations. While auxiliary sources provide cost-effective guidance in early iterations, our acquisition function also continues to select the target source throughout the optimisation.

\section{Details on Benchmark Problems}\label{ch:details_on_benchmarks}
We evaluate our method on several well-known constrained test problems to stress different aspects of performance:
\begin{itemize}
	\item The \textbf{Pressure Vessel} benchmark was proposed by \citet{coello_coello_constraint-handling_2002} with dimensionality $d=4$ and number of constraints $g=4$. The aim is to minimise a total cost of designing the pressure vessel, including the shell and head thicknesses, as well as the inner radius and length of the cylindrical section including some bounds.
	\item The benchmarks \textbf{Different Powers}, \textbf{Rastrigin} and \textbf{Rotated Rastrigin} (all $d=40$ and $g=9$) were drawn from the BBOB-constrained suite \citep{dufosse2022building}, These functions are constructed by applying non-linear transformations (e.g.\ rotations, asymmetric scaling, oscillatory distortions) to the base functions, then overlaying constraint functions so that the feasible region becomes non-trivial. For instance, the Rastrigin and rotated Rastrigin variants combine the highly multimodal base with global rotations, while Different Powers applies coordinate-wise power scaling to create differing levels of smoothness along axes. This set covers both multimodal, non-separable and ill-conditioned landscapes with constraints, making it a rigorous benchmark for constrained multi-source BO.
	\item Additionally, we test our method on the Rosenbrock objective \citep{rosenbrock_automatic_1960} augmented with two nonlinear constraints ($d=100$ and $g=2$) defined in \citet{eriksson_scalable_2021}, which challenges the model in high dimensions and with narrow feasible channels.
\end{itemize}

\section{Ablation and Parameter Study}\label{ch:ablation_and_param_study}
\begin{figure}[ht!]
	\centering
	\includegraphics[width=\textwidth]{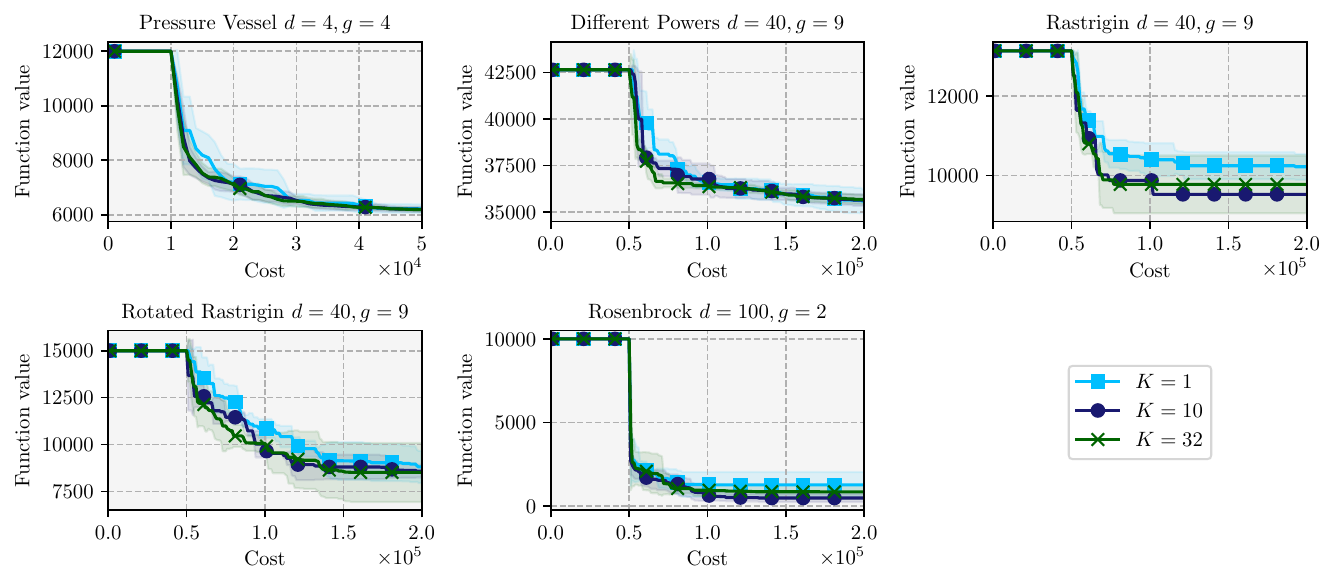}
	\caption{Parameter study on the number of MC samples $K$: optimisation performance for $K \in \{1, 10, 30\}$, showing that already moderate values yield stable results.}
	\label{fig:plot_k_param_study}
\end{figure}

\begin{figure*}
	\centering
	\includegraphics[width=\textwidth]{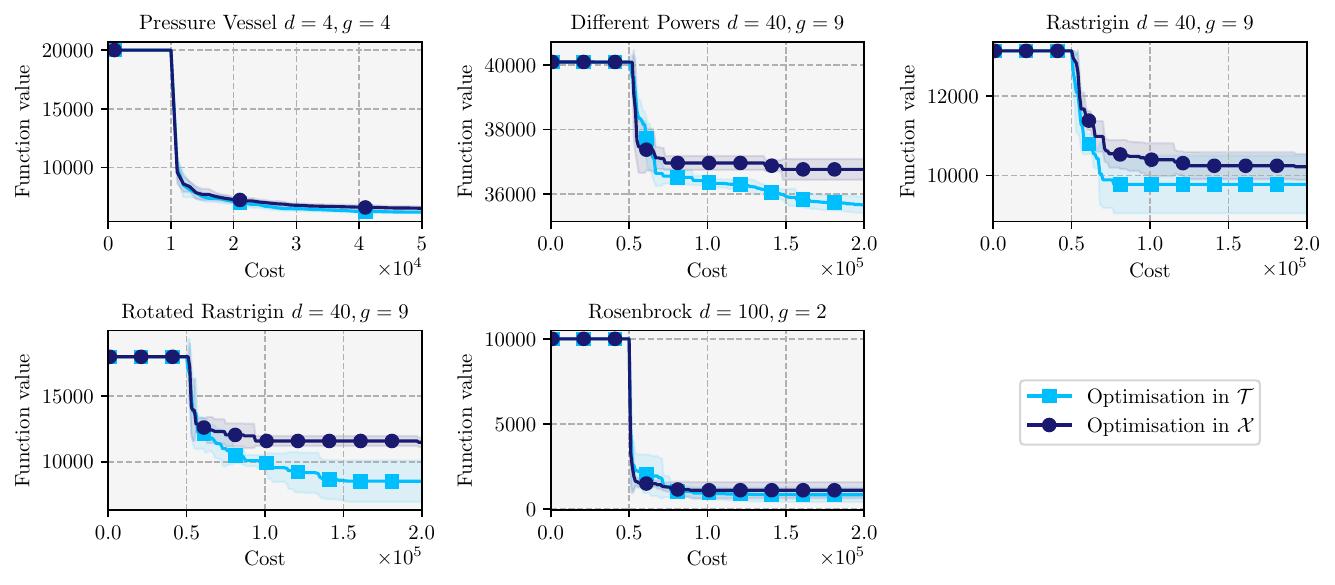}
	\caption{Ablation study on the trust region heuristic: comparison of MS-CMES with and without trust region–constrained acquisition optimisation.}
	\label{fig:tr_ablation}
\end{figure*}

\end{document}